\documentclass[conference]{IEEEtran}

\makeatletter

\def\ps@IEEEtitlepagestyle{%
  \def\@oddfoot{\mycopyrightnotice}%
  \def\@evenfoot{}%
}
\def\mycopyrightnotice{%
  {\footnotesize XXX-X-XXXX-XXXX-X/XX/\$XX.00~\copyright~20XX IEEE\hfill}
  \gdef\mycopyrightnotice{}
}

\usepackage{blindtext}
\usepackage{eso-pic}
\IEEEoverridecommandlockouts
\usepackage{cite}
\usepackage{amsmath,amssymb,amsfonts}
\usepackage{algorithmic}
\usepackage{graphicx}
\usepackage{textcomp}
\usepackage{xcolor}
\usepackage{tensor}
\usepackage{url}
\usepackage{subfigure}
\usepackage{hyperref}
\def\BibTeX{{\rm B\kern-.05em{\sc i\kern-.025em b}\kern-.08em
    T\kern-.1667em\lower.7ex\hbox{E}\kern-.125emX}}
    
\usepackage{eso-pic}
\newcommand\AtPageUpperMyright[1]{\AtPageUpperLeft{%
 \put(\LenToUnit{0.17\paperwidth},\LenToUnit{-2cm}){%
     \parbox{0.9\textwidth}{\raggedleft\fontsize{8}{11}\selectfont #1}}%
 }}%
\newcommand{\conf}[1]{%
\AddToShipoutPictureBG*{%
\AtPageUpperMyright{#1}
}
}

\begin{document}
\title{\vspace*{1cm} The Generalized Proximity Forest\\
}


\author{\IEEEauthorblockN{Ben Shaw}
\IEEEauthorblockA{\textit{Dept. of Mathematics \& Statistics} \\
\textit{Utah State University}\\
Logan, Utah, USA \\
ben.shaw@usu.edu}
\hspace{2.2in}
\and
\IEEEauthorblockN{Adam Rustad}
\IEEEauthorblockA{\textit{Dept. of Computer Science} \\
\textit{Brigham Young University}\\
Provo, UT, USA \\
arusty@byu.edu}
\hspace{2.2in}
\and
\IEEEauthorblockN{Sofia Pelagalli Maia}
\IEEEauthorblockA{\textit{Dept. of Statistics} \\
\textit{Brigham Young University}\\
Provo, UT, USA \\
smaia13@byu.edu}
\hspace{2.2in}
\and
\IEEEauthorblockN{Jake S. Rhodes}
\hspace{3.5in}
\IEEEauthorblockA{\textit{Dept. of Statistics} \\
\textit{Brigham Young University}\\
Provo, UT, USA \\
rhodes@stat.byu.edu}
\and
\IEEEauthorblockN{Kevin R. Moon}
\IEEEauthorblockA{\textit{Dept. of Mathematics \& Statistics} \\
\textit{Utah State University}\\
Logan, UT, USA \\
kevin.moon@usu.edu}
}

\maketitle
\conf{\textit{  Proc. of International Conference on Artificial Intelligence, Computer, Data Sciences and Applications (ACDSA 2026) \\ 
5-7 February 2026, Boracay-Philippines}}
\begin{abstract}
Recent work has demonstrated the utility of Random Forest (RF) proximities for various supervised machine learning tasks, including outlier detection, missing data imputation, and visualization. However, the utility of the RF proximities depends upon the success of the RF model, which itself is not the ideal model in all contexts. RF proximities have recently been extended to time series by means of the distance-based Proximity Forest (PF) model, among others, affording time series analysis with the benefits of RF proximities. In this work, we introduce the generalized PF model, thereby extending RF proximities to all contexts in which supervised distance-based machine learning can occur. Additionally, we introduce a variant of the PF model for regression tasks. We also introduce the notion of using the generalized PF model as a meta-learning framework, extending supervised imputation capability to any pre-trained classifier. We experimentally demonstrate the unique advantages of the generalized PF model compared with both the RF model and the $k$-nearest neighbors model.

\end{abstract}


\begin{IEEEkeywords}
classification, regression, proximities, imputation, outlier detection, ensemble methods, meta-learning
\end{IEEEkeywords}

\section{Introduction}
Random Forest (RF) proximities have proven useful for many applications, such as applications to financial data analysis, \cite{Li1}. However, the utility of RF proximities is limited to cases in which the RF model can be applied--namely, for tabular data. Recently, Geometry- and Accuracy-Preserving (GAP) RF proximities \cite{RFGAP} have been computed for time series models which exhibit a forest structure, such as the Proximity Forest (PF) model \cite{PF}, resulting in supervised outlier detection (of entire time series) \cite{PFGAP} and supervised missing data imputation for time series \cite{rhodes2025}. This recent work raises the important question of how to apply GAP proximities in the presence of additional data types, such as graph-valued and/or other cross-sectional data.

Pairwise distances can be considered for many types of data beyond tabular data: for example, Dynamic Time Warping (DTW) for time series data \cite{Comparison}, the Weisfeiler-Lehman distance for graph-valued data \cite{wldist}, and STRAMSim for portfolio (cross-sectional) data \cite{Li1}. Meanwhile, the PF model is a distance-based model, with splits determined not by features, as is the case with the RF model, but rather by proximity to chosen exemplars based on time series distances \cite{PF}. However, despite the fact that the original model was designed for univariate time series of equal length, the model logic is independent of the distance chosen. In this work, we consider the \textit{generalized proximity forest} model,
which modifies the original PF model in that it allows for the specification of custom distances in order to match the associated data, be it graph-valued, multivariate time series of unequal length, or ordinary tabular data. In so doing, we introduce GAP proximities to data domains not previously considered--any data domain, that is, for which pairwise distances may be considered.

In addition to the specification or selection of distances, we also expand the PF model in two novel ways. First, we devise a modified algorithm for the case of regression tasks, as the original PF model accommodates only classification tasks. Second, we devise a model-informed distance, endowing the generalized PF model with the status of a meta-algorithm, used primarily for imputation informed by any pretrained model: that is, we endow any pretrained model with the functionality of an imputer using GAP-based imputation.

This paper is organized as follows. In Section \ref{sec:background}, we provide background information for understanding our paper, including an overview of the PF model (\ref{sec:PF}), GAP proximities (\ref{sec:RFGAP}), and GAP-based imputation (\ref{sec:GAPimpute}). We discuss our methods in Section \ref{sec:methods}, which contains an overview of the functionality of the generalized PF model (\ref{sec:GenPF}) and a presentation on meta-learning for supervised imputation (\ref{sec:meta})
. Section \ref{sec:experiments} contains our varied experiments, which experiments demonstrate the functionality of the generalized PF model as well as its advantages over existing methods. We then conclude with Section \ref{sec:conclusion}. The source code, together with scripts for experiments, is available online.\footnote{\href{https://github.com/KevinMoonLab/PF-GAP}{https://github.com/KevinMoonLab/PF-GAP}, \href{https://github.com/KevinMoonLab/PF-GAP/tree/Manifold}{``manifold'' branch} for results.
}

\section{Background}
\label{sec:background}
In this section, we will present background information for our work. Section \ref{sec:PF} contains an overview of the PF model, and Section \ref{sec:RFGAP} gives an overview of RF proximities known as RF-GAP. Lastly, Section \ref{sec:GAPimpute} provides an overview of GAP-based imputation.

\subsection{Proximity Forests}
\label{sec:PF}

The Proximity Forest (PF) model was designed to be a scalable and accurate time series classifier \cite{PF}. One  motivation for the PF model was that standard RF model does not generally perform well on raw time series (using values at individual time points as features). This is due to the many ways in which a time series can be trivially perturbed, which perturbations cause the features fed into an RF model to be drastically different than the unperturbed time series~\cite{Goehry2023RFforTS}.

The PF model is similar to the RF model in that it is an ensemble of trees. In the PF model, the trees are "proximity trees," which are summarized as follows \cite{PF}. At each node, the number of branches created corresponds to the number of class labels present in the node. A random data sample, known as an "exemplar," is chosen for each class. Subsequently, the remaining data instances travel down the branch according to whichever exemplar they are most similar to, based on a pre-specified notion of distance. This step is repeated $r$ times, where $r$ is a hyperparameter, and the best of the $r$ outcomes is chosen according to a given notion of child node purity such as the Gini purity, which we use. This process is repeated until purity in the child nodes is obtained.

The original PF model was designed to accommodate 9 time series distances, which distances are randomly selected at each node or for each tree, according to the user selection \cite{PF}. Our modification of this algorithm gives a user full control over the distances used including the option of a single distance, allowing the PF algorithm to be applied more generally. 

Although KNN is another distance-based algorithm, one advantage the generalized PF model exhibits is its $\mathcal{O}(\log(N))$ inference complexity \cite{PF} compared with KNN's $\mathcal{O}(N)$ complexity, where $N$ is the number of training points. Although there are many acceleration methods for KNN that can reduce this complexity, such as ball \cite{omohundro1989balltree} and cover trees  \cite{beygelzimer2006covertree}, these methods typically rely on assumptions about the distance measure used, particularly the triangle inequality: we note, however, that the triangle inequality does not hold for many widely-used distance measures, including Dynamic Time warping and the Cosine distance.

\subsection{Geometry- and Accuracy-Preserving (GAP) Proximities}
\label{sec:RFGAP}

The development of a notion of tree-based similarity through random forest models was original work from Leo Breiman in the early 2000's~\cite{breiman2001rf-online}. The general idea was to define similarity between two points according to colocation in terminal nodes, which define the decision space of random forests. To explicitly define similarity, the number of terminal nodes in which a pair of points, $x_i$ and $x_j$, colocate was totaled, and this sum was normalized by the total number of trees trained. This notion was further extended in~\cite{RFGAP}, where it was determined that the aggregation of the total number of training points in a terminal node shared with an out-of-sample point, $x_i$, could define the similarity in such a way that the weighted sum of these similarities could perfectly reproduce the out-of-sample prediction of the random forest. Thus, these proximities were called random forest geometry- and accuracy-preserving proximities (RF-GAP). It has been demonstrated that the use of these RF-GAP proximities generally improved the performance of proximity-based applications such as data visualization, outlier detection, or missing value imputation, over other tree-based proximity measures~\cite{RFGAP, rhodes2023supervised-manifold}. 

The formal definition of RF-GAP proximities is as follows: given points $x_i$ and $x_j$, the GAP similarity between them is
\[
p(x_i, x_j) \;=\; \frac{1}{|S_i|} \sum_{t \in S_i} \frac{I\!\left(x_j \in J_i(t)\right)\, c_j(t)}{|M_i(t)|}.
\]
Here, $S_i$ denotes the set of trees in which $x_i$ is out-of-sample (e.g., not used to train the trees). For each tree $t \in S_i$, the multiset $J_i(t)$ contains the indices of in-bag observations whose points fall into the same terminal node as $x_i$. The quantity $c_j(t)$ records the number of times $x_j$ appears in the in-bag training sample for tree $t$, and $M_i(t)$ denotes the multiset of all in-bag indices associated with points that share the same terminal node as $x_i$, including bootstrap multiplicities. The function $I(\cdot)$ is the 0-1 indicator function. 

The definition was  adapted for  time-series data by slightly altering the formulation of proximity forests~\cite{PF} in \cite{PFGAP}. In this work, the authors introduced bootstrap resampling into the proximity forest framework. This modification was required to define out-of-sample sets, to enable the construction of GAP-style proximities. With this adjustment, the PF-GAP proximities are computed analogously to RF-GAP, using the ensemble of proximity trees for time series.  

The resulting proximities retain the key theoretical properties of RF-GAP. In particular, they satisfy proximity-weighted classification, ensuring that the out-of-sample prediction of the proximity forest can be reconstructed by a weighted-majority vote where the weights are given by the PF-GAP proximities. 

\subsection{GAP-based imputation}
\label{sec:GAPimpute}

The presentation of GAP-based imputation was originally given in \cite{rhodes2025} for the purpose of multivariate time series imputation, though we note that the algorithm naturally extends to similarly structured data, such as vector-valued data.

Let $\mathcal{D} = \{(x_n, y_n)\}_{n=1}^N$ denote a time series dataset of $N$ instances, where each instance $x_n \in \mathbb{R}^{p \times T}$ is a multivariate time series with $p$ features observed over $T$ time points, and $y_n \in \mathcal{Y}$ is the corresponding target label. For the purposes of this paper, $p = 1$ corresponds to univariate time series, although the methodology readily extends to $p > 1$ for multivariate settings for models capable of handling multivariate time series.

The full dataset is represented by a 3-dimensional array $X \in \mathbb{R}^{N \times p \times T}$, where the $(n, j, t)$-th entry, denoted $x_{njt}$, corresponds to the value of feature $j$ at time $t$ for instance $n$.

For each instance $n$ and feature $j \in \{1, \dots, p\}$, define the index sets of missing and observed time points:
\[
\mathcal{M}_{nj} = \{t : x_{njt} \text{ is missing} \}, \hspace{0.5em} 
\mathcal{O}_{nj} = \{t : x_{njt} \text{ is observed} \},
\]
so that $\mathcal{M}_{nj} \cup \mathcal{O}_{nj} = \{1, \dots, T\}$ for all $n, j$. Denote missing values as $x_{njt}^{\text{miss}}$ and observed values as $x_{njt}^{\text{obs}}$, with their imputations written as $\hat{x}_{njt}^{\text{miss}}$ and $\hat{x}_{njt}^{\text{obs}}$, respectively.

The GAP imputation procedure begins with an initial imputation of missing values in $X$ using basic strategies such as time-wise mean, median, or $k$-nearest neighbors ($k$-NN). This initialization can be performed globally across all time points or conditioned on the label $y_n$.

Once the imputed array, $\hat{X}$, is obtained, a time series classification model is trained on $(\hat{X}, y)$. From this model, GAP proximities $p(n, k)$ are derived for each pair of instances, quantifying similarity between point indices $n$ and $k$, as described in Section~\ref{sec:RFGAP}. These proximities act as adaptive weights for imputation, linking missing entries to similar observed ones.

Imputation proceeds iteratively over features and time points. For a given missing entry $x_{njt}^{\text{miss}}$ with $t \in \mathcal{M}_{nj}$, the imputed value is computed using observed values at $s \in \mathcal{O}_{nj}$ and proximity weights. For continuous features, we use a proximity-weighted average:

\[
\hat{x}_{njt}^{\text{miss}} = \sum_{k : t \in \mathcal{O}_{kj}} p(n, k) \, x_{kjt}.
\]


For categorical features with class set $\mathcal{C}_j$, we use a weighted majority vote:

\[
\hat{x}_{njt}^{\text{miss}} = \arg\max_{c \in \mathcal{C}_j} \sum_{k : t \in \mathcal{O}_{nj}} p(n, k) \cdot \mathbf{1}(x_{kjt} = c).
\]

To internally validate the imputation process, the imputation is applied to observed values $x_{njt}^{\text{obs}}$, treating them as pseudo-missing. This enables internal performance monitoring, analogous to out-of-bag evaluation in random forests. After one full pass updating all entries (both missing and pseudo-missing), a new model is trained on the updated $\hat{X}$, proximities are recomputed, and the process is repeated. The algorithm runs for a fixed number of iterations (typically 5), and the imputation yielding the best internal metric is selected.

The selection criterion is based on the reconstruction quality of re-imputed observed entries $\hat{x}_{njt}^{\text{obs}}$, using metrics such as $R^2$ for continuous features or $F_1$ score for categorical features. Alternatives like RMSE, MAE, or accuracy may also be specified.

For test-set imputation, given a new time series dataset $X^{\text{test}} \in \mathbb{R}^{N_{\text{test}} \times p \times T_{\text{test}}}$, the same initialization procedure is applied (without conditioning on labels), followed by extension of the trained GAP model to compute proximities between training and test instances. We assume that test instances are aligned over the same $T_{\text{test}}$ time points, leaving the extension to variable-length sequences for future work. Each missing entry $x_{njt}^{\text{test}}$ is then imputed using observed training data in feature $j$, weighted by the proximities $p^{\text{test}}(n, k)$ between test and training instances. In this way, test-set imputations benefit from label-informed structure learned during training without requiring labeled test data.

\section{Methods}
\label{sec:methods}

\subsection{The Generalized Proximity Forest Model}
\label{sec:GenPF}

As explained in Section \ref{sec:PF}, the PF model was originally designed as a (univariate) time series classifier \cite{PF}. The use of the Euclidean distance prohibits its use on time series of variable lengths, and the original implementation did not accommodate multivariate time series. In this paper, we propose the use of the PF model more generally, so that it can be used in whichever contexts KNN can: one needs only at least one notion of distance. As discussed in Section \ref{sec:PF}, this results in a model with lower asymptotic computational complexity than KNN--at least in cases where accelerations of KNN do not apply. Another benefit realized by the generalized PF model is in the extension of Random Forest proximities \cite{RFGAP, PFGAP} for data types lacking native support from the RF model. This extends the useful features of GAP-induced outlier detection and imputation to other data types.

Besides the stated conceptual contribution, we have also provided a flexible implementation of the PF model that accepts more generic data types. Importantly, users can not only specify the collection of distances used in the model, but also define their own distance functions in either Java (preferred), Python, or Maple. Missing data imputation with various initial methods, as well as obtaining outlier scores, are supported by making use of the GAP proximity graphs made available through the modified PF model \cite{PFGAP}.

We also adapt the PF model for regression tasks. Much of the PF algorithm is the same, though the way in which splits are handled changes substantially: each tree is strictly binary, and the randomly-sampled exemplars are drawn from the common pool of data instances per node. Node purity notions also change, with variance and mean absolute deviation being available choices in our implementation.

\subsection{Meta-learning for Supervised Imputation}
\label{sec:meta}

Although the PF model exhibited relatively strong performance as a time series classifier when it was introduced \cite{PF}, its performance was largely surpassed by subsequent models, including variants of the ROCKET model \cite{ROCKET}. However, many other models do not contain the benefit of GAP proximities, particularly for supervised imputation. To this end, we introduce the notion of using the generalized PF model so as to equip any pretrained classifier with imputation capabilities.

Let $f:\mathbb{R}^n \to Y$ be a pretrained classifier, where $Y$ is the space in which the distinct classes reside. For $x, y \in \mathbb{R}^n$, we define the following distance measure:
\begin{equation}{\label{meta_class}}
    d(x,y) = \begin{cases}
        0 & \text{  if } f(x) = f(y)\\
        1 & \text{  otherwise.}
    \end{cases}
\end{equation}
The distance measure in equation (\ref{meta_class}) is not a true metric, since $d(x,y)$ can be zero even when $x \ne y$. It is, however, symmetric, and it satisfies the triangle inequality. Alternatively, when a model outputs a vector $z \in \mathbb{R}^m$ of class probabilities, so that $f:\mathbb{R}^n \to \mathbb{R}^m$, we can define an $f$-informed distance between $x$ and $y$ simply by the Euclidean distance between $f(x)$ and $f(y)$.

Such a distance allows a pretrained model to be used to impute data as follows. First, one trains a PF model on the data used to train the pretrained model, or else on a representative subsample, using a model-informed distance such as the distance given in (\ref{meta_class}).

\section{Experiments}
\label{sec:experiments}


We begin our experiments by demonstrating the unique aspects of GAP proximities that are unavailable to the KNN model: in Section \ref{sec:penguin}, we demonstrate that GAP outlier scores are qualitatively different than KNN outliers, and the experiment in Section \ref{sec:2sphere} reinforces the notion that GAP-based imputation generally leads to better post-imputation accuracy than KNN-based imputation. We then turn to experiments for which the RF model cannot natively apply, namely graph classification (Section \ref{sec:graphexp}), and multivariate time series of unequal length (Section \ref{sec:jap}), comparing accuracies to KNN. In Section \ref{sec:meta_exp}, we demonstrate our proposed use of the generalized PF model as a meta-imputer, and Section \ref{sec:regression} showcases the PF model on a regression task. We conclude our experiments in Section \ref{sec:compare} with an experiment that compares the relative performance of KNN, the RF model, and the generalized PF model on small, vector-valued datasets.

\subsection{Palmer Penguin GAP Outliers}
\label{sec:penguin}

Having extended the PF model to more general data types, it is natural to consider the performance of the model in a simple use case relative to KNN. Thus, this experiment uses the Palmer Penguin dataset~\cite{horst2020palmerpenguins}, where the Euclidean distance is used for both the PF and KNN model.

Prior to training each model, we remove all rows with any missing data, leaving a total of 333 observations across seven variables. All variables are standardized to have a mean of zero with unit variance. For validation, separate training and test splits were randomly selected in respective proportions of 80\% and 20\%.

For a total of 10 independent trials with different train/test split initializations, a PF and a KNN model are trained. Each PF model has 11 trees with $r=5$, and each KNN model has $k=5$. Across all 10 trials, the mean difference between the obtained KNN accuracies and the obtained PF accuracies is $0.0044 \pm 0.0137$, where the error is represented by the standard deviation of the differences. Thus, the KNN and PF models are comparable, being statistically tied for performance.

\begin{figure*}[htp]
  \centering
  \subfigure{\includegraphics[width=.45\linewidth]{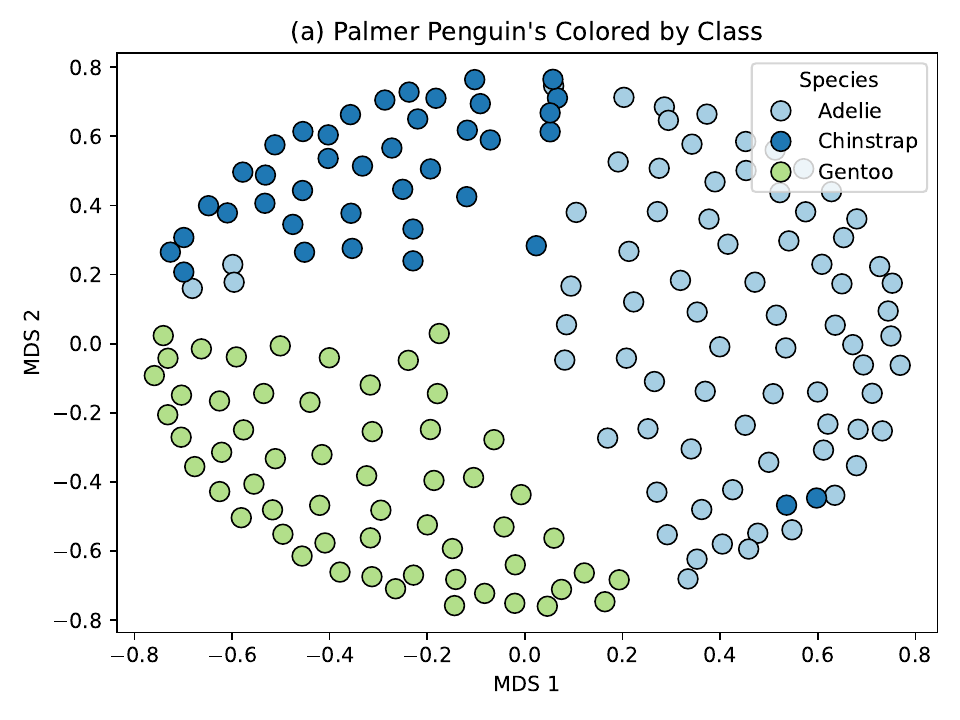}}\qquad
  \subfigure{\includegraphics[width=.45\linewidth]{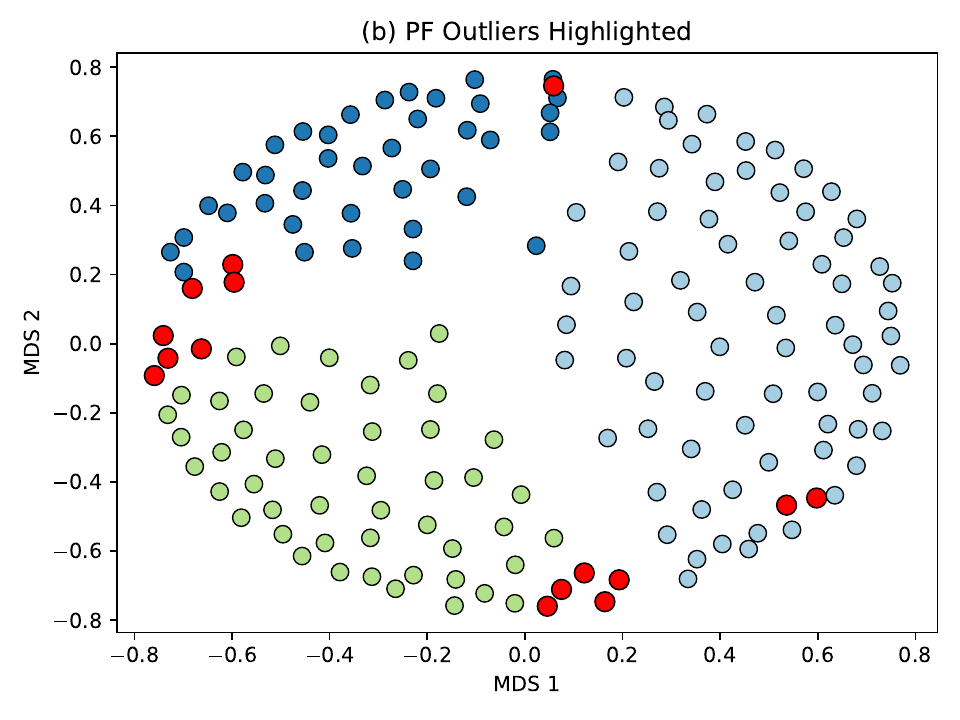}} 
  \caption{Left: MDS embedding of the palmer penguin dataset using PFGAP proximities. Right: the same embedding with points having the highest outlier scores for each class highlighted in red.}
  \label{fig:palmerpenguin}
\end{figure*}

In Figure \ref{fig:palmerpenguin}, we obtain an MDS embedding of the training points for a separate iteration which uses a train/test split of $50/50$. The distances used to obtain the MDS embeddings are obtained using the PF proximities, as described in Section~\ref{sec:RFGAP}. This embedding largely separates the classes, and we highlight the points with high outlier scores in red, which points often correspond to visual outliers in the MDS embedding.

\subsection{Missing Data Imputation on the 2-sphere}
\label{sec:2sphere}

In this binary classification experiment, we work with simulated data on the 2-sphere obtained using the \textit{geomstats} package for Python \cite{geomstats}. The dataset consists of two partially overlapping clusters on the unit sphere, generated by first sampling from a von Mises–Fisher distribution and then applying independent random rotations to produce two class-specific groups. For model validation, we use a train/test split of $50/50$.

The object of this experiment is not merely to compare classification accuracies of KNN and PF, but rather to compare post-imputation accuracies in the presence of missing data. Therefore, after generating our dataset, we remove (completely at random) $50\%$ of the data entries in both the train and test set, though ensuring that each data point has at least one feature still present. The standard Euclidean distance (in the latent space) is used for both models.

The KNN model imputes missing data using a proximity-weighted sum, where proximities are computed using the to-be-used KNN distance, ignoring missing values--that is, only considering distances along dimensions in which the two points in question have no missing data. GAP-based imputation proceeds as described in Section \ref{sec:GAPimpute}. 

The advantage of GAP-based imputation is seen in this experiment. The post-imputation test accuracy using KNN is found to be $0.8467$. For the PF model, the post-imputation test accuracy is found to be $0.8800$. Thus, this experiment highlights one potential advantage of the PF model over KNN: GAP-based imputation.

\subsection{Distance-based Graph Classification}
\label{sec:graphexp}

To illustrate that the generalized PF model can be applied to more generic data types, we consider a binary graph classification problem. The data for this experiment originates from the widely used \textit{TUDataset}, originally designed to benchmark methods for distinguishing enzymes (59\%) from non-enzymes (41\%). To ensure the classification task was non-trivial, the proteins were carefully selected so that no chain in the dataset showed significant structural similarity to any other outside its own parent structure~\cite{dobson2003proteins}.\footnote{This dataset is available as part of the \textit{PyTorch Geometric} package at: \url{https://pytorch-geometric.readthedocs.io/en/latest/generated/torch_geometric.datasets.TUDataset.html}.} We focus on the ``PROTEINS'' dataset, which provides $1113$ labeled graphs. We randomly sample $10\%$ for use as a validation set and another $10\%$ of the data is taken to be the test dataset.

For both the KNN and PF models, we use the Weisfeiler–Lehman distance \cite{wldist}. For KNN, we use $k=5$, and for PF we use 11 trees. The PF model obtains validation and test accuracies of $0.7027$ and $0.7768$, respectively. On the other hand, the validation and test accuracies for KNN are, respectively, $0.5405$ and $0.6250$. This demonstrates that the relative performance of the KNN and PF models, distance-for-distance, can be task-dependent. 


\subsection{Japanese Vowels: Accuracy test of KNN vs PF}
\label{sec:jap}

In this experiment, we turn to the benchmark dataset knows as JapaneseVowels \cite{MV}. In this dataset, each instance is a multivariate time series, and each time series does not necessarily have the same length. Thus, this time series dataset is beyond the reach of the original PF model, which was designed for univariate time series data of equal length \cite{PF}.

We compare our generalized PF model with an implementation of KNN using \textit{aeon} \cite{aeon} with Dynamic Time Warping. Using the provided train/test split of the data, we obtain an accuracy score of $0.9486$ with $k=1$, which accuracy does not substantially change with $k=5$. Using the PF model with independent and dependent Dynamic Time Warping, we obtain an accuracy score of $0.9757$. Due to the non-deterministic qualities of the PF model, we run the experiment using PF repeatedly, obtaining scores in $[0.96, 0.98]$.


While previous experiments suggest that the KNN and PF model are comparable in a ``distance-for-distance'' manner, this experiment demonstrates that the PF model can have a clear and decisive advantage over KNN in terms of model accuracy.

\subsection{Meta Imputation with the Mini-ROCKET model}
\label{sec:meta_exp}

We turn now to an experiment that demonstrates the ability to use the generalized PF model as a meta algorithm for imputation. We look to the \textit{ArrowHead} dataset from the 2018 UCR Archive \cite{ucr}, as this is a dataset for which the MiniROCKET model, a variant of the ROCKET model \cite{ROCKET, minirocket}, seems to outperform the original PF model in terms of test accuracy. We train the MiniROCKET model using the provided training set, obtaining a model accuracy of approximately $0.8629$. We then train a PF model on the training set, using the distance defined in (\ref{meta_class}). Then, artificially creating a "missing completely at random" test dataset with $10\%$ missing data, we perform GAP-based imputation using the trained PF model (initializing with linear imputation), obtaining a PF test accuracy score of $0.8686$ using the class matching distance defined by the pretrained MiniROCKET model. This experiment suggests that the generalized PF model can be coupled with pretrained models, enabling the pretrained models to be used as an advanced imputer before final model prediction.

\subsection{Regression with PF}
\label{sec:regression}

In our last experiment, we turn to the \textit{FloodModeling1} dataset \cite{tser}, which is a univariate time series regression task. Using KNN with $k=5$ with dynamic time warping yields a test $R^2$ score of $0.7949$. The PF model, with $100$ trees, dynamic time warping distance, and the mean absolute deviation serving as node purity, obtains a test $R^2$ score of $0.9068$. This result demonstrates the potential of the generalized PF model for regression tasks. 

\subsection{Euclidean Distances for Vector-valued data}
\label{sec:compare}

\begin{figure}
    \centering
    \includegraphics[width=0.95\linewidth]{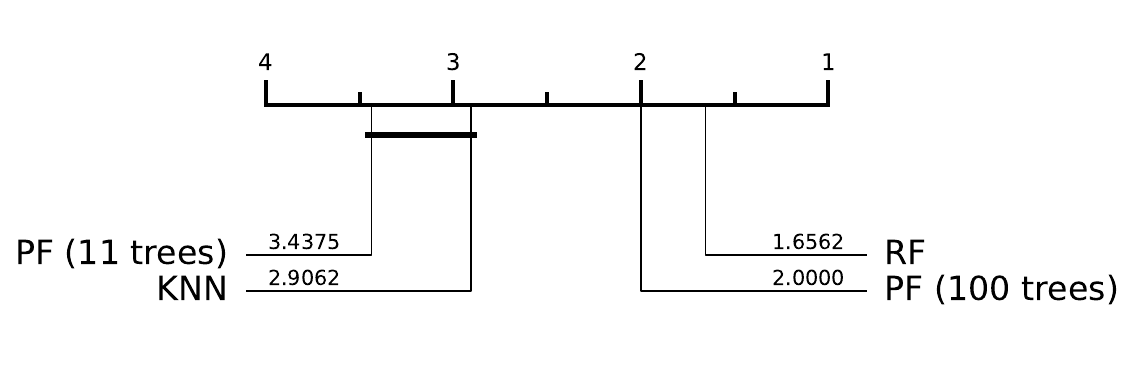}
    \caption{Critical difference plot of the RF, PF, and KNN models in 31 small, vector-valued datasets. The RF model usually ranks the best, followed by the PF model with 100 trees. The KNN model and the PF model with 11 trees are statistically tied.}
    \label{fig:cd}
\end{figure}

Having introduced the notion of using the PF model in more general settings, it is natural to consider the relative performance of the PF model in simple settings. We therefore compare the generalized PF model, using Euclidean distance only, to KNN with Euclidean distance and the RF model on 31 datasets available from the OpenML-CC18 Classification benchmark, including all datasets limited to 5000 observations and 100 features \cite{oml-benchmarking-suites}. For each dataset, the accuracy for each model is recorded 10 times. Using the means for each dataset, a critical difference diagram is shown in Figure \ref{fig:cd}.

The results in Figure \ref{fig:cd} were obtained using a $k$-value of $5$ for KNN. For the RF model, $100$ trees were used. For the PF model, two separate approaches were utilized: the first with $11$ trees, and the second with $100$ trees, both using $r=5$. 

This experiment suggests that the generalized PF model is comparable to KNN for small, vector-valued datasets, though the RF model appears to dominate in terms of test accuracy. Subsequent experiments will show that GAP proximities present a compelling reason to use the PF model instead of the KNN model: however, this result demonstrates that, for certain domains of interest, the RF model may be better suited.

\section{Conclusion}
\label{sec:conclusion}
We have generalized the previously-given classifier known as Proximity Forests (PF) to obtain a distance-based model for extending Random Forest (RF) proximities to data types beyond time series which cannot natively be handled by the RF model. We have shown that the generalized PF model tends to compare with $k$-nearest neighbors (KNN) in terms of model performance while also highlighting the advantages offered by the PF model. We have shown that GAP proximities are substantially different than KNN proximities, reinforcing the notion that GAP-based imputation tends to yield higher post-imputation classification accuracy than KNN-based imputation, and we have highlighted the fact that the PF model scales more favorably than the brute-force KNN algorithm.

In addition to realizing the PF model with data types beyond univariate time series of equal length, we have also introduced a modified PF algorithm adapted for regression tasks. This adds crucial functionality, as the RF and KNN models are adapted to regression tasks.

We have also introduced the notion of using the generalized PF model as a meta algorithm for imputation using pretrained models. In this way, any pretrained model can now be used as a supervised imputer using the GAP-based algorithm. This allows for supervised imputation to be accomplished potentially more successfully if certain models outperform current models making use of GAP proximities.

For future work, we seek to apply the generalized PF model to additional data types, such as text data, in a future endeavor. We also intend to consider the generalized PF model alongside other forest-based models for model performance and GAP-based performance in outlier detection and post-imputation scores.

\section*{Acknowledgment}

We wish to acknowledge those who have worked and continue to work to create and maintain the useful time series benchmark datasets contained in the UCR \cite{ucr} and TSER \cite{tser} archives. These useful datasets better enable development of machine learning techniques.

\bibliographystyle{ieeetr}
\bibliography{main}


\end{document}